# *FireFly*: Autonomous Rescue Drones

Hajer Ben Mnaouer, Mohammad Faieq, Adel Yousefi, Sarra Ben Mnaouer, *Canadian University Dubai*, E-mail: *20190007782@students.cud.ac.ae*, *20190008013@students.cud.ac.ae*
*20180006563@students.cud.ac.ae*, *20180007044@students.cud.ac.ae*,
*Adel Ben Mnaouer, Omar Mashaal,* *adel@cud.ac.ae*, *omar.mashaal@cud.ac.ae*
*First Interchange Sheikh Zaid Road, Dubai, United Arab Emirates.*

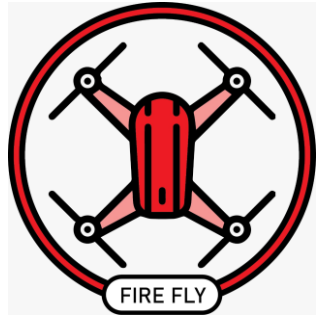

**Abstract**

As a fire erupts, the first few minutes can be critical, and first respondents must race to the scene to analyze the situation and act fast before it gets out of hand. Factors such as road traffic condition and distance may not allow quick rescue operation using traditional means and methods, leading to unmanageable spreading of fire, injuries or even deaths that can be avoided. *FireFly* drone-based rescue consists of a squad of highly equipped drones that will be the first responders to the fire site. Their intervention will make the task of the fire rescue team much more effective and will contribute to reduce the overall damage. As soon as the fire is detected by in-building implanted sensors, the fire department would deploy a set of *FireFly* drones that would fly to the site, scan the building, and send live fire status information to the Fire fighter team. The drones would have the ability to identify trapped humans using AI based pattern recognition tools (using sensors and thermal cameras) and then drop them rescue kits as appropriate. The drones will also be equipped with fire detection and recognition capabilities and be able to drop fire extinguishing balls as first attempts to put off seeds of fires before they evolve. The integration of drones with firefighting will allow for ease of access and control of fire outbreaks. Drones will also result in increased response time, prevention of further damage, and allow relaying of vital information to out of reach places regarding the characteristics of the fire scene.

## 1. INTRODUCTION

The combat against fire outbreaks in the UAE has proven to be a prevalent issue in the past years. Being a first world developed country, the UAE is famous for its high towers and skyscrapers, but with that advancement comes great risks as fires can be big hazards when dealing with such high-rise buildings. On New Year's Eve of 2016, The Address hotel was set ablaze at 9:30 pm. The fire originated on the 20$^{th}$ floor of the building and due to the height of its location, it proved to be difficult to localize the flames and get them under control. Moreover, due to the extensive traffics of Dubai's busy New Year's Eve roads and streets, fire fighters and rescue teams had a difficult time attempting to reach the ablaze hotel. In 2017, in Marina Dubai the torch tower (ironically) caught flames as well a second time in a row in two years, first time catching fire back in 2015. These past fire occurrences highlight the importance of immediate response in such situations, and how the first few minutes of the fire can be so critical. Those extra minutes can cost human lives, more injuries and ruined architecture leading to great losses.

## 2. PROBLEM

The issue with combating fire stems from one critical point, time. As a fire ignites, the flames may spread quickly, and those first few moments can be quite critical. Firefighting tools have always been low-tech such as trucks, ladders and hoses which does not always allow for quick extinguish time. Having a delayed response can further worsen the situation and sometimes lead it to uncontrollable state of affairs. Another critical hurdle consists in the skyscrapers' heights at which some of these fires may erupt. With conventional firefighting equipment it can be impossible to reach high story buildings and therefore any attempts to curb the fire will involve fire fighters be futile. Furthermore, the lack of information at the fire scene can lead to firefighters' inability to containing the fire. Information such as how extensive the fire has spread and how many potential victims may be trapped. All that plays an important role into the issues that are dealt with when a fire erupts.

## 3. PROPOSED SOLUTION

Considering the issues that arise when dealing with fire combat, the proposed solution is a fire rescue drone. According to [1] a study by Goldman Sachs estimates that there will be an $881 million market for drone firefighting by 2020, with exponential annual growth. The percentage is unknown, but it is expected to rise in the US and Europe. These drones will be equipped with all the necessary parts to help fight fires more efficiently. The drones can be categorized according to specific tasks and purposes. One drone can be responsible for extinguishing the fire. These drones would be equipped with fire extinguishing balls formed from hard foam shell which burst with instant contact with flames and disperse an ABC dry chemical powder which helps extinguish the fire. They will also include thermal imaging cameras that allow firefighters to identify hotspots and search for unsafe locations in the building that require fire extinguishing. They would also include sensors and a trained AI that can detect and predict the spread of the fire. Another drone can have the responsibility of breaking large windows of the building and providing trapped victims with fire rescue kits. These rescue kits can include fire masks, fire blankets, first aid kits, and oxygen masks. The drone will have a camera with facial recognition and added AI features will help recognize a person's condition and can also help lead victims to exits of the building. One more drone can be responsible for figuring out vital information about the fire, such as the extent of the fire, the level of air toxicity, the nature of fire that has erupted. The advancements that come with using firefighting and rescue drones such as the cameras, sensors and battery life are expected to help firefighters reach higher ground and extinguish fires more efficiently.

## 4. BENEFITS

- Early fire containment
- Increased chance of rescue
- Better access to fire in high areas
- Faster arrival at the scene
- Acquiring more data and information about the fire
- Better understanding of the scene
- Decreasing damage costs in the long run
- Increase chance of victim survival and early fire control
- Faster access to the medical kits for trapped victims
- Reducing rescue costs
- Monitoring the census of the victims
- Increased safety to the firefighters

## 5. FUNCTIONALITY AND SPECIFICATIONS OF DRONE

The drone will have a fly time of 45 minutes. 20 minutes can be dedicated to and from the drone station allotting the remaining 25 min for fire control and victim rescue. After its return to the charging station, the drone can go through a smart battery replacement station. This mechanism can be done in mere seconds saving valuable time. The battery/charging station will be powered by solar panels. The drone will be equipped with a gyroscope to have a balanced flight and added GPS to help navigate to the exact location of the fire. The drone will communicate with the ground station using LoRa (Long Range) technology which is a proprietary low-power wide-area network modulation technique

The drone has a modular design which enables it to perform 4 main functions which will help control and maintain the fire.

First, to help locate the locations of the fire, the drone will scan the building using the dedicated camera system composed of a FLIR thermal imaging camera, infrared heat cameras to provide night vision, and a 3D radar camera to provide a 3d map and plot the most heated areas. Upon detection of the fire, the drone's emergency hammer feature will be employed to break through tough glass and gain access into the inside of the building for better rescue ability. Then the combination of these cameras will help distinguish and provide a clear understanding of where the fire is located and how extensive it has spread. The data gathered by the camera system will be processed using Artificial Intelligence to provide further information about how the fire may have spread.

The AI system will also help predict which locations will be best for the second main function that is control of the fire using a special cannon that deploys fire extinguishing balls. These special balls are formed from hard foam shell and burst with instant contact of the flames and disperse an ABC dry chemical powder that will help extinguish the fire. The cannon system will use two high speed motors which would shoot the balls with a high but safe velocity. The fire extinguishing balls are shot inside the building into the most fire dense areas (determined by the AI system) and that will help provide the most optimal fire combat and allow for better control of the fire. Onboard the drone will be a ***jetson nano***, a small, powerful computer that can run multiple neural networks in parallel for image classification and object detection applications.

Thirdly, the drones will be equipped with vision sensors, facial recognition, image recognition, motion detection and thermal cameras and sensors that will aid in the identification of the accurate location of the fire victims. The image recognition can help recognize nearest exit signs. The drone will also include a high beam light which will help victims recognize that the drone is coming towards them and help them visualize their way through the dark or smoke. Fire fighters will be able to watch the scene thru video streaming from the drone back to the ground station and will also be speak to the victims through the speaker in the drone and attempt to guide them to safety. Furthermore, firefighters will be able to voice control the drones and remotely coordinate their actions thus overriding autonomous operations when needed.

The Gas sensors can help distinguish which type of fire has erupted and, in that way, help combat it accordingly.

Fires can be classified into:
- Class A - fires involving solid materials such as wood, paper or textiles.
- Class B - fires involving flammable liquids such as petrol, diesel or oils.
- Class C - fires involving gases.
- Class D - fires involving metals.
- Class E - fires involving live electrical apparatus.
- Class F - fires involving cooking oils such as in deep-fat fryers. [2]

|  | CLASS A | CLASS B | CLASS C | CLASS D | CLASS E | CLASS F |
|---|---|---|---|---|---|---|
| AFF FOAM | ✓ | ✓ |  |  |  |  |
| WATER | ✓ |  |  |  |  |  |
| WET CHEMICAL | ✓ |  |  |  |  | ✓ |
| CO2 |  | ✓ |  |  | ✓ |  |
| DRY POWDER | ✓ | ✓ | ✓ |  | ✓ |  |
| L2 & M28 POWDER |  |  |  | ✓ |  |  |

**Fig. 1:** Classification of Fires and appropriate extinguishers [2]

So, depending on which fire class has been detected by the gas sensors, the fire fighters can then decide what is the immediate course of action that is most suitable for the situation.

The fourth main function is that the drones will be able to administer emergency fire kits that will be shot out of the cannons or dropped to reach the victims. These fire kits will include fire masks, fire blankets, first aid kits, and oxygen masks to help increase their chance of survival.

All the data that is gathered will be sent back to the fire department in real time and accessed using the smart screen and tablets. These tablets are supposed to be deployed in building corridors and will be gathering sensors data and will be the first devices to detect fire eruption and send instant notifications to firefighting departments.

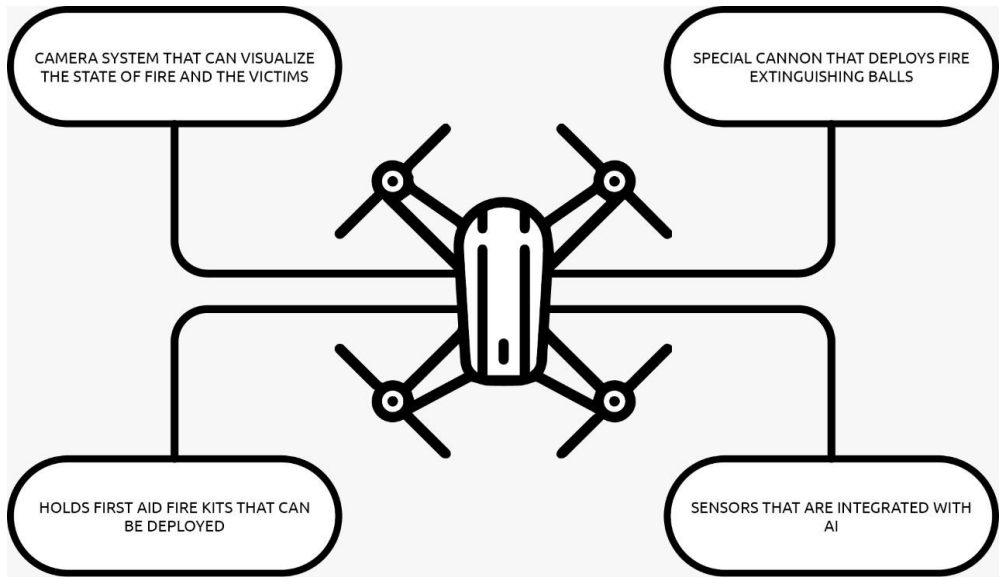

**Fig. 2:** Four Functions of the *FireFly* Drone.

## 6. COMPONENTS AND TECHNOLOGY

### 6.1. Camera (gateway and receiver)
- Proposed camera option is FLIR camera, a thermal imaging camera which are special fire fighting cameras that we can integrate in the drone system.
- Infrared heat cameras to provide night vision
- 3D radar camera to provide a 3D map and plot the most heated areas

### 6.2. Sensors
- Gas sensor
- Drone Body temperature Sensor
- Obstacle detection – Laser radar sensor
- Motion detection

### 6.3. Lights
- Powerful LED lights to light up the dark areas

### 6.4. Brain
- PX4 autopilot with Raspberry Pi 4
- Lora Antenna
- GPS
- Flight control

### 6.5. Speaker

### 6.6. Payload
- Able to carry a minimum load of 5 Kgs
- Canon that shoots (breaks window) (shoots/drops rescue kits)
- Carry and shoot fire extinguishing balls

### 6.7. Connected to smart screen/Tablet
- Able to display camera information / heat maps
- Send commands from smart screen to drones/firefighting depts.
- Receiving live sensor and visual feedback

**6.8. Tablets that receive the information from sensors, analyze and decide about fire occurrence. Notify the firefighting dept and trigger drone rescue missions.**

**6.9. Battery (with solar power charging station) / energy harvesting capability (if possible)**

## 7. COSTS

The estimated costs of the drone will be 30,000-50,000 AED in the early stages of prototyping. When the product reaches the manufacturing stage the price of the product will rapidly decrease.

## 8. INVESTORS AND STAKEHOLDERS

### 8.1. Investors:

The intended investors of the project would be the government but would also include private investors interested in aiding such an initiative. They would be interested in the general security and prevention of fire outbreaks in their buildings and local properties and therefore they would be willing to fund the project.

The potential investors for this project could be: -
- The Government – Be it for incidents related to hazardous materials, motor vehicle accidents around town, wildfires, or even rescue operations; the firefly drone could attract huge amounts of investments from the government that solely prioritize the safety of their countries. *FireFly* would be also a substantial contribution to making the city a smart one.
- Incubators – Business incubators assist startups to easily access resources such as financing on their business venture. Given the innovation and importance within the project, many business incubators are sure to invest in FireFly.
- Military & Defense – while the government could use the drone within the country, the military and defense could be additional potential investors for the project as the drone guarantees safety at the borders.
- Industrial Operations – fires are common amongst factories where industrial operations are in progress. The FireFly is a huge opportunity for such corporates to reduce their hefty costs and avoid disruptions to work. As a result, companies tied to Industrial Operations could perhaps be the biggest investors from the corporate world.
- Hospitality Industry – The hospitality industry guarantees utmost comfort and peace to its guests. As a result, many hoteliers would be keen on investing in FireFly as the massive production and distribution of the drones would add an extra star to the services they provide to their customers.

### 8.2. Stakeholders:

Certain individuals and/or organizations have a vested interest in drones like *FireFly* and hence can either affect their business' operations and performance or be affected by it. Outlining these stakeholders is key to ensuring no business activity negatively impacts those that are affected by the product.
1. Owners/Drone Manufacturers: Be it the producers of the drone or owners of the brand under which they are provided, such individuals would be perhaps the biggest stakeholder for FireFly.
2. Fire Organizations: Many countries possess both public fire brigades and private fire protection organizations – both of these are of interest to this project as they have a vested interest in the drones.
3. Government: the government can highly influence the business activities and consequences of this project since it is crucial for the country's safety and will push forward city smartness.
4. Insurance companies :insurance companies would like to decrease the damage costs resulting from fires and other casualties.
5. Municipality and civilians: aiding to secure the local and public areas around the country.
6. Customers of private fire protection organizations – though these individuals might not have a high interest in the business activities of such drones, they still do impact the adoption of drones in the society by supporting or constraining the products.

## 9. CONCLUSION

This paper proposes building a conceptual system prototype to help combat fire using drones as first responders. Traditional methods of firefighting are still relatively low-tech, and the use of ladders and water hoses is not enough to help resolve the fire as efficiently as possible. Integration of Drones with fire combat will help with quicker response, easier access to high-rise buildings, added information about the fire scene, vital data about victims' count and incident environment, and a general improvement to the firefighting task. The drones will have four main functionalities that include various components. A camera system composed of thermal and infrared visualizations for sending live time video shooting to the firefighting department, a shooting mechanism to launch fire extinguishing balls to control the fire, different types of sensors that will help identify the location of the fire victims and send back information about the type of fire and gasses being emitted and will be able to provide emergency fire kits help aid the victims. The future of fire combat will be greatly improved with the incorporation of autonomously operated drones that are AI/ML empowered making them exactly what is needed to solve most of the issues and setbacks that fire fighters experience. Their incorporation in several monitoring missions is a hot trend in making cities smarter.